\documentclass[runningheads]{llncs}
\usepackage{graphicx}
\usepackage{algorithm} 
\usepackage{algpseudocode}
\usepackage{tikz}
\usepackage{comment}
\usepackage{amsmath,amssymb} 
\usepackage{color}
\usepackage{kotex-logo}
\usepackage{kotex}
\usepackage{multirow}
\usepackage{wrapfig}
\usepackage{array}
\usepackage[accsupp]{axessibility}  

\begin{document}
\pagestyle{headings}
\mainmatter
\def\ECCVSubNumber{7648}  

\title{Hyperparameter Optimization \\with Neural Network Pruning} 


\titlerunning{HPO with NNP}
%
\author{KANG IL LEE\inst{1} \and
JUN HO YIM\inst{2}}
\authorrunning{KANG IL LEE, JUN HO YIM}
%
\institute{AIRS Company, Hyundai Motor Group \\
    \email{kangil-lee@hyundai.com \inst{1}} \email{junho.yim@hyundai.com \inst{2}}
}
\maketitle

\begin{abstract}
\label{abs}
Since the deep learning model is highly dependent on hyperparameters, hyperparameter optimization is essential in developing deep learning model-based applications, even if it takes a long time. As service development using deep learning models has gradually become competitive, many developers highly demand rapid hyperparameter optimization algorithms. In order to keep pace with the needs of faster hyperparameter optimization algorithms, researchers are focusing on improving the speed of hyperparameter optimization algorithm. \\
\indent However, the huge time consumption of hyperparameter optimization due to the high computational cost of the deep learning model itself has not been dealt with in-depth. Like using surrogate model in Bayesian optimization, to solve this problem, it is necessary to consider proxy model for a neural network~($N_{\mathcal{B}}$) to be used for hyperparameter optimization.
Inspired by the main goal of neural network pruning, i.e., high computational cost reduction and performance preservation, we presumed that the neural network~($N_{\mathcal{P}}$) obtained through neural network pruning would be a good proxy model of $N_{\mathcal{B}}$. In order to verify our idea, we performed extensive experiments by using CIFAR10, CFIAR100, and TinyImageNet datasets and three generally-used neural networks and three representative hyperparameter optmization methods. Through these experiments, we verified that $N_{\mathcal{P}}$ can be a good proxy model of $N_{\mathcal{B}}$ for rapid hyperparameter optimization. The proposed hyperparameter optimization framework can reduce the amount of time up to 37\%.

\keywords{Hyperparameter Optimization, Neural Network Pruning}
\end{abstract}

\section{Introduction}
\label{Intro}
\begin{figure}[t]\centering
        \begin{center}
            \includegraphics[trim=0.0cm 0.0cm 0.0cm 0.0cm,width=\columnwidth]{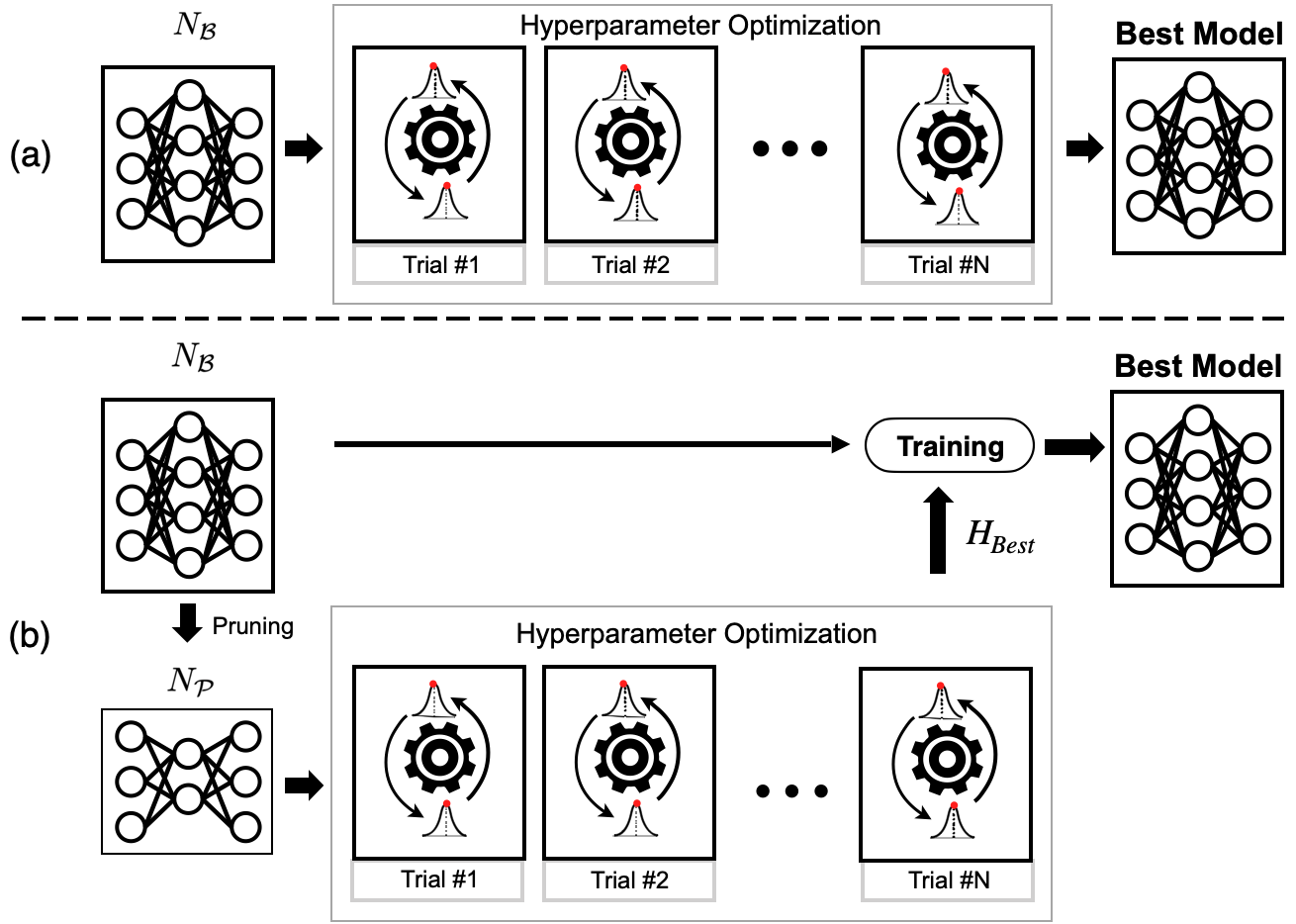}
        \end{center}
        \caption{Comparison between (a) typical hyperparameter optimization and (b) hyperparameter optimization with neural network pruning. While (a) utilizes a target neural network~($N_{\mathcal{B}}$) directly, (b) uses a pruned neural network~($N_{\mathcal{P}}$). After finding best hyperparameters~($H_{Best}$), (b) uses $H_{Best}$ to train $N_{\mathcal{B}}$.}
        \label{fig:HPO Procedures}
    \end{figure}

Recently, as application development using deep learning models has become more active, there is an increasing demand for a way to provide models quickly. Hyperparameter optimization~(HPO) is one of the factors hindering rapid development, but it is one of the inevitable steps to obtain good performance. Therefore many researchers are still manually tuning hyperparameters based on a their own experience. This is a very tedious task and it is difficult to guarantee optimal performance. In order to avoid this tedious task, many researchers have been studying HPO for a long time~\cite{bergstra2011algorithms,bergstra2013making,wistuba2015hyperparameter,li2017hyperband,balandat2020botorch}.\\
\indent HPO research can be divided into three categories according to the optimization method: random search, evolutionary optimization, Bayesian optimization. Random search is a method that performs random sampling on the probability distributions of hyperparameters until a predetermined budget is exhausted or the desired performance is achieved. Evolutionary optimization considers multiple hyperparameter sets sampled from probability distributions as one generation and leaves only a strong hyperparameter set among them. Then, it aims to find the best probability distributions by repeatedly performing the process of updating the parameters (e.g. mean and variance) of probability distributions using the remaining set. Bayesian optimization-based HPO~(BOHPO) estimates the probability distribution of an objective function and tries to find the most promising hyperparameters in the estimated distribution. In Bayesian optimization, the Gaussian Process is used as a probabilistic surrogate model because direct estimation of the probability distribution of the objective function requires a high amount of computation. Recent HPO research focuses on BOHPO because BOHPO has more scalability than the other HPO methods~\cite{balandat2020botorch,ha2019bayesian,kandasamy2020tuning,cakmak2020bayesian}. In addition, hyperparameter pruning that reduce the search space to perform HPO quickly and accurately is continuously being studied~\cite{wistuba2015hyperparameter,li2017hyperband,jamieson2016non}.\\
\indent However, despite the advance of HPO research, it is still difficult to find the optimal hyperparameter set using limited resources due to the high computational cost caused by the deep learning model itself. Therefore, it is necessary to consider what can replace the target neural network, just as the Gaussian process is used as a surrogate model during Bayesian optimization. Neural network pruning aims to minimize the performance gap between the neural network~($N_{\mathcal{P}}$) obtained by pruning and the corresponding base neural network~($N_{\mathcal{B}}$) and to reduce the amount of computation as much as possible. Inspired by the goal of neural network pruning, we presumed that $N_{\mathcal{P}}$ can be used as a good proxy model of $N_{\mathcal{B}}$. As described in Fig.\ref{fig:HPO Procedures}.(b), prune a neural network. Next, perform HPO by using $N_{\mathcal{P}}$. Lastly, train $N_{\mathcal{B}}$ with the best hyperparameters obtained by HPO using $N_{\mathcal{P}}$. We confirmed the feasibility of this idea through a simple experiment~(See Table \ref{tbl:motivation}). The performance of $N_{\mathcal{B}}$ obtained through Fig.\ref{fig:HPO Procedures}.(b) was comparable to that of $N_{\mathcal{B}}$ obtained through Fig.\ref{fig:HPO Procedures}.(a)~(See Table \ref{tbl:motivation}). Based on this result, we performed extensive experiments using various datasets, models, and HPO methods to verify whether the observation could be generalized. In the consequence of experiments, we verified that $N_{\mathcal{P}}$ can be a good proxy model of $N_{\mathcal{B}}$. Thus, we propose to use $N_{\mathcal{P}}$ as a proxy model of $N_{\mathcal{B}}$ for rapid HPO. The proposed HPO framework can reduce the amount of time up to 37\%, and it can be applied regardless of HPO algorithm, type of neural network, and dataset.

\section{Related Works}
\label{related}
\indent \textbf{Single Shot Pruning.}~Since Lecun et al.~\cite{lecun1990optimal} introduced neural network pruning as a solution to the over-parameterization problem, many researchers have actively studied neural network pruning until nowadays. Recently, \textit{single shot pruning} methods were proposed~\cite{lee2018snip,wang2020picking}. Lee et al.~\cite{lee2018snip} proposed an unstructured pruning method at the initialization state based on the connection sensitivity score~(SNIP). Lee et al.~\cite{lee2018snip} aimed to avoid the heuristic hyperparameter decisions of pruning and iterative pruning procedures requiring a high computational cost. Wang et al.~\cite{wang2020picking} improved the pruning criteria of SNIP by using Hessian. On the other hand, Tanaka et al.~\cite{tanaka2020pruning} pointed out that when a high compression ratio is used to prune a neural network, an entire layer of a neural network could be pruned~(\textit{layer collapse}). To solve \textit{layer collapse}, Tanaka et al.~\cite{tanaka2020pruning} proposed an iterative pruning method with an exponential pruning schedule. Meanwhile, van et al.~\cite{van2020single} extended SNIP and proposed a structured pruning method at initialization state~(3SP). At first, in the 3SP algorithm, calculate the gradient over loss function for all weights. After that, calculate FLOPs of all layers. The FLOPs of all layers are used to adjust channel pruning scores. This approach may try to avoid \textit{layer collapse} by adjusting channel scores through the calculated FLOPs. \\ 
\indent Our variation of Synflow~\cite{tanaka2020pruning} is different in the perspective of handling \textit{layer collapse} and scoring a channel. To avoid \textit{layer collapse}, our variation remains minimum channel which allow a model to learn data properly according to predefined minimum remaining channel ratio. Since our variation assigns score for individual weights in first step based on Synflow and differently calculate a channel score, it results in a pruned neural network far from 3SP's one. \\
\indent \textbf{Hyper Parameter Optimization.}~HPO is one of the important parts of AutoML and has been studied for a long time since the 1990s when machine learning applications began to gain popularity. The purpose of the study on HPO is that ordinary people or most researchers get to be free from the hyperparameter searches. HPO research started from grid search. Grid search divides a search space into $n$ compartments to find a hyperparameter based on human knowledge. It can be better than manual search in terms of time efficiency. But in the worst case, if we have $k$ hyperparameters, it has a complexity of~$O(n^{k})$, moreover, unless the search space is set up precisely, the possibility of achieving high performance is low. Random search predetermines the number of trials and randomly selects a hyperparameter from a probability distribution until converges to the desired performance or runs out of budget. Although random search has higher possibility to obtain the highest performance than grid search within budget, random search still needs expert knowledge for search space to find globally optimal hyperparameter. HPO methods based on Bayesian optimization~(BOHPO) have been proposed to address the shortcomings of the previous HPO methods~\cite{snoek2012practical,springenberg2016bayesian,bergstra2013making}. BOHPO mitigates the scalability problem of the preceding three methods through \textit{exploitation, exploration} evaluating the vicinity and far areas respectively from the currently selected hyperparameter set. Roughly speaking, BOHPO repeats two processes alternately. First, sample hyperparameters through the acquisition function, and surrogate function evaluates the hyperparameters. After that, repeat two processes until it converges to a predefined value or reaches optimum. \\
\indent The mainly covered hyperparameters in HPO research can be classified into various groups by the characteristics of values such as continuous~(e.g.,learning rate), discrete~(e.g.,batch size), and categorical~(e.g.,optimizer)~\cite{yang2020hyperparameter}. We conducted experiments using representative of random search, evolutionary optimization, and Bayesian optimization to confirm the generality of our observation~\cite{bergstra2011algorithms,hansen2016cma,balandat2020botorch}. And among the hyperparameters that have been mainly dealt with in HPO research, most experiments were conducted by focusing on the hyperparameters that researchers tune most frequently, such as learning rate, batch size, and weight decay.

\section{Hyperparameter Transfer}
\label{HPO}
Since most deep learning models take high computational costs in the forward and backward passes, HPO requires considerable time for a single search process. For this reason, despite the advance of hyperparameter optimization research, there are still many cases where the hyperparameters of deep learning models are manually tuned depending on empirical evidence. Therefore, search space pruning methods have been continuously proposed to alleviate this problem~\cite{wistuba2015hyperparameter,jamieson2016non,li2017hyperband}. Nevertheless, it still takes a lot of time to find an optimal hyperparamters. \\
\indent To solve this problem, we need a proxy model with a low computational cost that can be used for HPO instead of the target neural network~($N_{\mathcal{B}}$). Neural network pruning aims at high computational cost reduction and performance conservation. Since this goal perfectly satisfies the conditions of the aforementioned proxy model, we presumed that the neural network~($N_{\mathcal{P}}$) obtained through pruning could be utilized as a good proxy model.
\begin{table}[!t]
    \centering
    \caption{Feasibility experiments. NM, NS, and 3SP mean pruning methods proposed by \cite{kim2020neuron}, \cite{liu2017learning}, and \cite{van2020single}, respectively. We used Hyperband~\cite{li2017hyperband} as search space pruner and TPE~\cite{bergstra2011algorithms} as hyperparameter optimizer. The last two columns show the accuracy of the test set of each dataset. Ours refers to the single shot pruning method which is transformed to structured pruning form from Synflow~\cite{tanaka2020pruning}~(See Section 4).}
    \label{tbl:motivation}
    {
        \begin{tabular}{c|c|c|cc}
        \hline
        \multirow{2}{*}{Pruning Method} & \multirow{2}{*}{Model} & \multirow{2}{*}{Dataset} & \multicolumn{2}{c}{Accuracy (\%)}      \\ \cline{4-5} 
                                  &                        &                          & \multicolumn{1}{c|}{Typical HPO} & With Pruning \\ \hline
        NM~\cite{kim2020neuron}   & ResNet56               & CIFAR10                  & \multicolumn{1}{c|}{93.43}    &  93.57 \\
        Ours                      & ResNet56               & CIFAR10                  & \multicolumn{1}{c|}{93.43}    &  93.62 \\ \hline
        NS~\cite{liu2017learning} & VGG19-BN               & CIFAR10                  & \multicolumn{1}{c|}{93.77}    &  92.15 \\ 
        Ours                      & VGG19-BN               & CIFAR10                  & \multicolumn{1}{c|}{93.77}    &  93.31 \\ \hline
        3SP~\cite{van2020single}  & ResNet56               & CIFAR100                 & \multicolumn{1}{c|}{71.24}    &  71.76 \\
        Ours  & ResNet56               & CIFAR100                 & \multicolumn{1}{c|}{71.24}    &  72.31 \\ \hline
        \end{tabular}
    }
\end{table}
The detail of our idea is follows. As it can be seen in Fig.\ref{fig:HPO Procedures}.(b), prune a neural network. Next, perform HPO by using $N_{\mathcal{P}}$. Lastly, train $N_{\mathcal{B}}$ with the best hyperparameters obtained by HPO using $N_{\mathcal{P}}$. In order to verify this idea, we performed experiments by using the three existing pruning methods~\cite{van2020single,kim2020neuron,liu2017learning} and the pruning method that we transformed into a structured pruning form of Synflow~\cite{tanaka2020pruning}~(See Table \ref{tbl:motivation}). When using the two pruning methods~\cite{liu2017learning,kim2020neuron} that require pretrained neural network in each experiment, HPO was performed after reinitialization of the weights of each pruned neural network. This is because, if the neural network obtained by pruning the pretrained neural network is used for HPO without reinitialization, it will be optimized for a hyperparameter suitable for fine-tuning the pruned neural network itself. Meanwhile, we transformed Synflow~\cite{tanaka2020pruning}, one of the single-shot pruning methods~\cite{lee2018snip,wang2020picking,tanaka2020pruning} that do not require a pretraining stage, into a structured pruning form and used it in the experiment~(See section \ref{Pruning}). We compared the performance of hyperparameters by measuring the top-1 accuracy on the validation set of each dataset.\\
\indent As shown in Table \ref{tbl:motivation}, hyperparameters optimized by using $N_{\mathcal{P}}$ as a proxy model of $N_{\mathcal{B}}$ showed comparable performance to those through typical HPO. In other words, hyperparameter transfer is possible from $N_{\mathcal{P}}$ to $N_{\mathcal{B}}$. The HPO process using pruning is as follows~(See Fig.\ref{fig:HPO Procedures}.(b)). First, prune base neural network~($N_{\mathcal{B}}$). Next, by using $N_{\mathcal{P}}$ for HPO, find a hyperparameter set~($\mathcal{H}_{\mathcal{P}}$). Lastly, the best model is obtained by using the acquired best hyperparameter~($\mathcal{H}_{Best}$) in the training of $N_{\mathcal{B}}$. Since hyperparameter transfer performance is almost independent of the pruning method, it is recommended to use the single-shot pruning-based method that requires a very small amount of computation.

\section{The Variation of Single Shot Pruning}
\label{Pruning}
Most of pruning methods generally take three steps~\cite{liu2017learning,huang2018data,he2019filter,wang2019eigendamage,meng2020pruning}. i) Train a neural network. ii) Prune the pretrained neural network through their pruning algorithm. iii) Finetune the pruned neural network. However, the recently proposed single-shot pruning methods do not require pretraining~\cite{lee2018snip,wang2020picking,tanaka2020pruning}. For example, SNIP~\cite{lee2018snip} put a single mini batch as an input in the initial state and prune using the gradients obtained from it without pretraining. Therefore, there is no cost required for pretraining. We presumed that the performance of hyperparameter transfer is independent of the pruning method depending on the results of Table \ref{tbl:motivation}. Thus, we decided to use single-shot pruning method for computational efficiency. \\
\indent However, since the previous single-shot pruning methods~\cite{lee2018snip,wang2020picking,tanaka2020pruning} are unstructured pruning requiring irregular operations to reduce practical computation cost, it is difficult to take advantage of pruning without a dedicated device or a library~\cite{han2016eie}. Therefore, we transformed Synflow~\cite{tanaka2020pruning} to structured pruning. In order to reduce the amount of computation as much as possible, we remove the iterative pruning process of Synflow, and even the part that changes the sign of the weight to unsigned in the pruning score calculation process. In addition, to obtain a score suitable for structured pruning, the average value was taken based on the fan-out axis, and this was used as the score for each channel. And \textit{layer collapse} was prevented by adding a simple channel conservation constraint. The details are as follows. The scores of individual weights are calculated as Eq.\ref{eq2}. 
\begin{equation} 
    \label{eq2}
    \begin{aligned}
        S_{\mathcal{W}}^{o,i,h,w}= \left | \frac{\partial L(\textbf{x})}{\partial \mathcal{W}_{o,i,h,w}} \odot \mathcal{W}_{o,i,h,w} \right |.
    \end{aligned}
\end{equation}
$\odot$~ means element wise product, \textbf{x} means input, and $\mathcal{W}_{o,i,h,w}$ means weight of 2D convolution filter. We used the sum of logits as the loss function $L$ according to the conservation law of synaptic saliency claimed from Synflow~\cite{tanaka2020pruning}. After calculating the individual score, the score for each channel is calculated as Eq.\ref{eq3}.
\begin{equation} \label{eq3}
    Score_{o}=\sum_{i}\sum_{h}\sum_{w} \frac{S_{\mathcal{W}}^{o,i,h,w}}{i \times h \times w},~ Score_{o} \in R^{o} .
\end{equation}
Then, each component of $Score_{o}$ is stored in global score set~($G_{scores}$). After calculating the scores for all channels, sort the $G_{scores}$ in ascending order and set the threshold according to the desired percentage~($\rho$). Then, prune the channels which have lower score than the threshold. If the ratio of the remaining channels of a layer is lower than the predefined minimum remaining channel ratio~($\sigma$), it is forced to remain additional channels which allow a layer to preserve the ratio of $\sigma$ compared to the number of original channels. This process prevents layer collapse. Meanwhile, in the case of the linear layer, it is assumed that there are no two dimensions corresponding to height and width in the explanation of the convolution layer above, and all calculations are performed identically. On the other hand, the normalization layer is excluded from the scoring process, and the remaining dimension is determined according to the front and rear layers. Algorithm \ref{alg:pruning alg} shows our pruning method and flow of applying the above description to the entire layer. More details about pruning, such as how to deal with residual connection, group convolution, and layer dependency, are described in the Supplementary material.
\begin{algorithm}[!h] 
	\caption{Pruning Algorithm} 
	\label{alg:pruning alg}
	\begin{algorithmic}[1]
	\small
	\State \textbf{Require:}~Set $\rho \in [0, 100]$ and $\sigma \in [0, 1]$
	\State \textbf{Initialize}~$G_{scores} \gets \emptyset $
	\State \textbf{Input:} A mini batch 
	\State \textbf{Output:} A pruned neural network
	\State \textbf{Step.1:}~Calculate the gradients of neural network for given input\\
	    \hspace{\algorithmicindent} $g_{\mathcal{W}_{o,i,h,w}} \gets \frac{\partial \mathcal{L}}{\partial \mathcal{W}_{o,i,h,w}}$
	\State \textbf{Step.2:}~Calculate the scores of individual weights\\
	    \hspace{\algorithmicindent} $S_{\mathcal{W}_{o,i,h,w}}=g_{\mathcal{W}_{o,i,h,w}}\odot\mathcal{W}_{o,i,h,w}$ 
	\State \textbf{Step.3:}~Calculate the score of each channel\\
	    \hspace{\algorithmicindent}     
	    $Score_{o}=\sum_{i}\sum_{h}\sum_{w} \frac{S_{\mathcal{W}}^{o,i,h,w}}{i \times h \times w},~ Score_{o} \in R^{o}$ \\
	    \hspace{\algorithmicindent} $G_{scores} \gets G_{scores}~\cup~Score_{o}$
	\State \textbf{Step.4:}~Prune the channels having lower score than threshold~($\tau$)\\
	    \hspace{\algorithmicindent} $\tau :$ $\rho$-th percentile of the channel scores\\
	    \hspace{\algorithmicindent} Let $N(\cdot):$ be the number of elements for an arbitrary set\\
	    \hspace{\algorithmicindent} \textbf{Initialize.} Group $\mathcal{G}
	    \gets \emptyset$ \\
	    \hspace{\algorithmicindent} Sort $G_{scores}$ in ascending order.
	    \For {$l$ in $layers$}
	        \State $C_{l}$ : The set of channels for $l$
	        \State \textbf{Initialize.} Group $C \gets \emptyset$ 
	        \For {$c~~in~~C_{l}$}
	            \If{$Score~of~c \geq \tau $}
	                \State $\mathcal{C} \gets \mathcal{C} \cup c$
	            \EndIf
	        \EndFor
    	    \If{$\frac{N(\mathcal{C})}{N(C_{l})} \geq \sigma$ }
    	        \State $\mathcal{G} \gets \mathcal{G} \cup C $
    	    \Else
    	        \State $\mathcal{C} \gets \mathcal{C} \cup \mathcal{C+}$
    	        \State $\mathcal{C+}$ : Additional channels to guarantee minimum channel remaining ratio.
    	        \State $\mathcal{G} \gets \mathcal{G} \cup \mathcal{C}$
    	    \EndIf
	    \EndFor
	    \State \Return $\mathcal{G}$
    \end{algorithmic}
\end{algorithm}

\section{Experiments}
\label{exp}
\subsection{Experimental Setup}
We used well-known three datasets:CIFAR10, CIFAR100~\cite{krizhevsky2009learning}, and TinyImageNet~\cite{Le2015TinyIV} and models : VGG16-BN~\cite{simonyan2014very}, ResNet56~\cite{he2016deep}, and MobileNetV2~\cite{sandler2018mobilenetv2} in various experiments. We used \textit{hyperband}~\cite{li2017hyperband} as search space pruner and 
TPE~\cite{bergstra2011algorithms}, BO~\cite{balandat2020botorch}, CMA-ES~\cite{loshchilov2016cma} as representative hyperparameter optimization methods. We used SGD optimizer with momentum(=0.9) and decayed learning rate at 80 epoch and 120 epoch by 10 times in most experiments. We mainly targeted three hyperparameters, which most researchers mainly adjust: learning rate, weight decay, and batch size. Additionally, we verified the scalability of our observation by adding the other two hyperparameters~(learning rate decay term, and SGD momentum). The search spaces of hyperparameters and allocated time budget are described in Table \ref{tbl:HPO_range}. We used the learning rate, weight decay, and batch size as the main target hyperparameters, and the learning rate decay term and SGD momentum were used only in additional experiments. In most experiments, we utilized open source hyperparameter optimization framework \textit{optuna}~\cite{optuna_2019}.\\
\begin{table}[h]
\centering
\caption{The target hyperparameters and Time Budget}
\label{tbl:HPO_range}
\begin{tabular}{c|ccccc|cc}
\hline
         & \multicolumn{5}{c|}{Hyperparameters}                                                                                                                                                                                                                                                                                                                                                     & \multicolumn{2}{c}{Time Budget}                                                    \\ \hline
Category & \multicolumn{1}{c|}{\begin{tabular}[c]{@{}c@{}}Learning \\ Rate\end{tabular}} & \multicolumn{1}{c|}{\begin{tabular}[c]{@{}c@{}}Batch \\ Size\end{tabular}} & \multicolumn{1}{c|}{\begin{tabular}[c]{@{}c@{}}Weight \\ Decay\end{tabular}} & \multicolumn{1}{c|}{\begin{tabular}[c]{@{}c@{}}Learning Rate \\ Decay\end{tabular}} & \begin{tabular}[c]{@{}c@{}}SGD\\ Momentum\end{tabular} & \multicolumn{1}{c|}{Epoch} & \begin{tabular}[c]{@{}c@{}}HPO \\ Trials\end{tabular} \\ \hline
Value    & \multicolumn{1}{c|}{1e-5 $\sim$2e-1}                                          & \multicolumn{1}{c|}{32$\sim$1024}                                          & \multicolumn{1}{c|}{1e-5$\sim$2e-2}                                          & \multicolumn{1}{c|}{1e-1$\sim$1}                                                    & 1e-1$\sim$9.9e-1                                       & \multicolumn{1}{c|}{160}   & 50                                                    \\ \hline
\end{tabular}
\vspace{1mm}
\end{table}
\indent We pruned neural networks with a high percentile~($\rho$) value 85 in most experiments to take sufficient reduction in computational amount. And we set minimum channel remaining ratio~($\sigma$) value 0.15 to guarantee normal training of each neural network. In terms of data augmentation, we used random horizontal flip~(probability=0.5) and random crop~(zero padding=4, size=32) for CIFAR Series. In the case of TinyImageNet, we used random horizontal flip~(probability=0.5), random rotation~(degree=$15^{\circ}$) and random-crop~(zero padding=6, size=128).
\begin{table}[!t]
\centering
\caption{Experimental results on whether the neural network obtained through pruning can be used as a good proxy model. `Time Reduction' means the averagely reduced percentage of the time required from the forward pass to the backward pass for the same batch size~(The higher is better).}
\label{tbl:main_results}
\begin{tabular}{c|c|c|cc|c}
\hline
\multirow{3}{*}{Dataset}      & \multirow{3}{*}{\begin{tabular}[c]{@{}c@{}}Hyperparameter\\ Optimizer\end{tabular}} & \multirow{3}{*}{\begin{tabular}[c]{@{}c@{}}Neural\\ Network\end{tabular}} & \multicolumn{2}{c|}{Accuracy~(\%)}                                                                                                                                            & \multirow{3}{*}{\begin{tabular}[c]{@{}c@{}}Time \\  Reduction\end{tabular}} \\ \cline{4-5}
                              &                                                                                   &                                                                           & \multicolumn{1}{c|}{\multirow{2}{*}{\begin{tabular}[c]{@{}c@{}}Typical\\ HPO\end{tabular}}} & \multirow{2}{*}{\begin{tabular}[c]{@{}c@{}}With   \\ Pruning\end{tabular}} &                                                                             \\
                              &                                                                                   &                                                                           & \multicolumn{1}{c|}{}                                                                       &                                                                            &                                                                             \\ \hline
\multirow{9}{*}{CIFAR10}      & \multirow{3}{*}{TPE}                                                              & ResNet56                                                                  & \multicolumn{1}{c|}{$93.43 \pm 0.23 $}                                                      & $93.62 \pm 0.32$                                                           & 12 \%                                                                         \\ \cline{3-6} 
                              &                                                                                   & VGG16-BN                                                                  & \multicolumn{1}{c|}{$93.41 \pm 0.42$}                                                                    & $93.61 \pm 0.07$                                                                        & 23 \%                                                                         \\ \cline{3-6} 
                              &                                                                                   & MobileNetV2                                                               & \multicolumn{1}{c|}{$94.54 \pm 0.14$}                                                                    & $94.87 \pm 0.36$                                                                        & 32 \%                                                                         \\ \cline{2-6} 
                              & \multirow{3}{*}{BO}                                                            & ResNet56                                                                  & \multicolumn{1}{c|}{$93.04 \pm 0.37$}                                                                    & $93.18 \pm 0.58$                                                                        & 12 \%                                                                         \\ \cline{3-6} 
                              &                                                                                   & VGG16-BN                                                                  & \multicolumn{1}{c|}{$93.70 \pm 0.14$}                                                                    & $93.94 \pm 0.22$                                                                        & 23 \%                                                                         \\ \cline{3-6} 
                              &                                                                                   & MobileNetV2                                                               & \multicolumn{1}{c|}{$94.58 \pm 0.42$}                                                                    & $95.27 \pm 0.10$                                                                        & 32 \%                                                                         \\ \cline{2-6} 
                              & \multirow{3}{*}{CMA-ES}                                                           & ResNet56                                                                  & \multicolumn{1}{c|}{$93.11 \pm 0.39$}                                                       & $93.50 \pm 0.20$                                                           & 12 \%                                                                         \\ \cline{3-6} 
                              &                                                                                   & VGG16-BN                                                                  & \multicolumn{1}{c|}{$93.30 \pm 0.36$}                                                                    & $93.79 \pm 0.21$                                                                        & 23 \%                                                                         \\ \cline{3-6} 
                              &                                                                                   & MobileNetV2                                                               & \multicolumn{1}{c|}{$94.66 \pm 0.65$}                                                                    & $95.40 \pm 0.10$                                                                        & 32 \%                                                                         \\ \hline
\multirow{9}{*}{CIFAR100}     & \multirow{3}{*}{TPE}                                                              & ResNet56                                                                  & \multicolumn{1}{c|}{$71.24 \pm 0.23$}                                                       & $72.31\pm 0.73$                                                            & 12 \%                                                                         \\ \cline{3-6} 
                              &                                                                                   & VGG16-BN                                                                  & \multicolumn{1}{c|}{$73.51 \pm 0.34 $}                                                      & $73.36 \pm 0.47$                                                           & 22 \%                                                                         \\ \cline{3-6} 
                              &                                                                                   & MobileNetV2                                                               & \multicolumn{1}{c|}{$77.94 \pm 2.22$}                                                                    & $77.89 \pm 1.47$                                                                        & 37 \%                                                                         \\ \cline{2-6} 
                              & \multirow{3}{*}{BO}                                                            & ResNet56                                                                  & \multicolumn{1}{c|}{$71.23 \pm 1.19$}                                                                    & $71.74 \pm 1.59$                                                                        & 12 \%                                                                         \\ \cline{3-6} 
                              &                                                                                   & VGG16-BN                                                                  & \multicolumn{1}{c|}{$73.00 \pm 0.60$}                                                                    & $73.43 \pm 0.34$                                                                        & 22 \%                                                                         \\ \cline{3-6} 
                              &                                                                                   & MobileNetV2                                                               & \multicolumn{1}{c|}{$76.20 \pm 1.66$}                                                                    & $78.57 \pm 0.72$                                                                        & 37 \%                                                                         \\ \cline{2-6} 
                              & \multirow{3}{*}{CMA-ES}                                                           & ResNet56                                                                  & \multicolumn{1}{c|}{$71.32 \pm 0.54$}                                                                    & $71.05 \pm 0.45$                                                                        & 12 \%                                                                         \\ \cline{3-6} 
                              &                                                                                   & VGG16-BN                                                                  & \multicolumn{1}{c|}{$72.25 \pm 0.42$}                                                                    & $73.47 \pm 0.30$                                                                        & 22 \%                                                                         \\ \cline{3-6} 
                              &                                                                                   & MobileNetV2                                                               & \multicolumn{1}{c|}{$74.78 \pm 1.52$}                                                                    & $75.90 \pm 1.03$                                                                        & 37 \%                                                                         \\ \hline
\multirow{9}{*}{TinyImageNet} & \multirow{3}{*}{TPE}                                                              & ResNet56                                                                  & \multicolumn{1}{c|}{$52.92 \pm 0.80$}                                                                    & $52.29 \pm 0.94$                                                                        & 20 \%                                                                         \\ \cline{3-6} 
                              &                                                                                   & VGG16-BN                                                                  & \multicolumn{1}{c|}{$61.49 \pm 0.55$}                                                                    & $61.03 \pm 0.44$                                                                        & 16 \%                                                                         \\ \cline{3-6} 
                              &                                                                                   & MobileNetV2                                                               & \multicolumn{1}{c|}{$57.99 \pm 0.44$}                                                                    & $56.99 \pm 1.40$                                                                        & 23 \%                                                                         \\ \cline{2-6} 
                              & \multirow{3}{*}{BO}                                                            & ResNet56                                                                  & \multicolumn{1}{c|}{$51.72 \pm 0.36$}                                                                    & $51.34 \pm 0.76$                                                                        & 20 \%                                                                         \\ \cline{3-6} 
                              &                                                                                   & VGG16-BN                                                                  & \multicolumn{1}{c|}{$56.48 \pm 2.25$}                                                                    & $58.44 \pm 1.96$                                                                        & 16 \%                                                                         \\ \cline{3-6} 
                              &                                                                                   & MobileNetV2                                                               & \multicolumn{1}{c|}{$57.84 \pm 0.42$}                                                                    & $56.08 \pm 1.45$                                                                        & 23 \%                                                                         \\ \cline{2-6} 
                              & \multirow{3}{*}{CMA-ES}                                                           & ResNet56                                                                  & \multicolumn{1}{c|}{$52.96 \pm 0.70$}                                                                    & $52.60 \pm 0.30$                                                                        & 20 \%                                                                         \\ \cline{3-6} 
                              &                                                                                   & VGG16-BN                                                                  & \multicolumn{1}{c|}{$56.27 \pm 2.27$}                                                                    & $58.78 \pm 2.01$                                                                        & 16 \%                                                                         \\ \cline{3-6} 
                              &                                                                                   & MobileNetV2                                                               & \multicolumn{1}{c|}{$57.87 \pm 1.72$}                                                                    & $56.32 \pm 1.82$                                                                        & 23 \%                                                                         \\ \hline
\end{tabular}
\end{table}
\begin{table}[!t]
\centering
\caption{Hyperparameter transferability experiment on ResNet Family. Each column represents the accuracy of ResNet56 for CIFAR100 validation set. The left side of $\rightarrow$ refers to the neural network performing HPO, and the right side refers to the target neural network for hyperparameter transfer.}
\label{tbl:family networks}
\begin{tabular}{c|c|c|c|c}
\hline
             & Baseline         & \begin{tabular}[c]{@{}c@{}}ResNet8\\  $\rightarrow$ ResNet56\end{tabular} & \begin{tabular}[c]{@{}c@{}}ResNet14\\  $\rightarrow$ ResNet56\end{tabular} & \begin{tabular}[c]{@{}c@{}}ResNet32\\  $\rightarrow$ ResNet56\end{tabular} \\ \hline
Accuracy(\%) & $72.31 \pm 0.73$ & $71.66 \pm 0.38$                                                       & $71.76 \pm 0.39$                                                        & $71.60 \pm 0.74$                                                        \\ \hline
\end{tabular}
\vspace{2mm}
\end{table}
\subsection{Hyperparameter Optimization with Pruning}
    We conducted experiments on commonly used three datasets and models to verify our presumption. In addition, we used three hyperparameter optimization algorithms: random search~\cite{bergstra2011algorithms}, evolutionary strategy~\cite{loshchilov2016cma}, and Bayesian optimization~\cite{balandat2020botorch} to prove that our presumption is independent of the hyperparameter optimizer. And we used \textit{Hyperband}~\cite{li2017hyperband} as a search space pruner for the overall experiments. All experiments corresponding to each row of the Table \ref{tbl:main_results} were performed using 5 random seeds. And we measured the accuracy for the validation set of each dataset to quantify the performance of hyperparameters found through HPO.\\
    \indent Table \ref{tbl:main_results} shows the performance of hyperparameter sets obtained through typical HPO framework~(Fig.\ref{fig:HPO Procedures}.(a)) and the proposed HPO framework which uses a pruned neural network as a proxy model~(Fig.\ref{fig:HPO Procedures}.(b)). We can see there are marginal gaps when comparing the case where the target neural network is directly used for HPO and the case where the proxy model is used for HPO. In other words, the performance of the hyperparameter sets found through each procedure is almostly equal. If the hyperparameter optimizer is different, even if the dataset and network are the same, there are performance differences. This phenomenon may be attributed to the performance difference of the hyperparameter optimizer. In terms of time reduction, the rate of reduction was lower than expected, due to the property of the neural network~($N_{\mathcal{P}}$) obtained through the pruning method used in experiments. Since the single-shot pruning method uses an initial weight, if we use weight initializations that are commonly used~\cite{he2015delving,glorot2010understanding}, relatively more channels of the low-level layer than high-level layer are inevitably left. This is because the weights will be initialized more sparsely as it goes to a higher-level layer, and the pruning score for each weight is calculated according to Eq.\ref{eq2}. As a result, since the channels of the low level layer are remained at a relatively high rate, the time reduction rate was lower than the actual number of parameters reduction rate.\\
    \indent Taken together, the experimental results are consistent with the trend of the Table \ref{tbl:motivation} that became our motivation. And this trend is independent of dataset, hyperparameter optimizer, and model. Therefore, we concluded that it is a good choice to use the neural network obtained through pruning as a proxy model to reduce the time required for hyperparameter optimization.
\subsection{Hyperparameter Transferability Analysis}
\indent In this section, we analyze the reason of why a pruned neural network can be used as a proxy model. And we discuss whether our proposal is valid even when conducting HPO for more hyperparameters. \\
\begin{wrapfigure}{r}{0.5\textwidth}{h}
  \begin{center}
    \vspace{-1.7cm}
    \includegraphics[trim=0.0cm 0.0cm 0.0cm 0.0cm,width=0.5\textwidth]{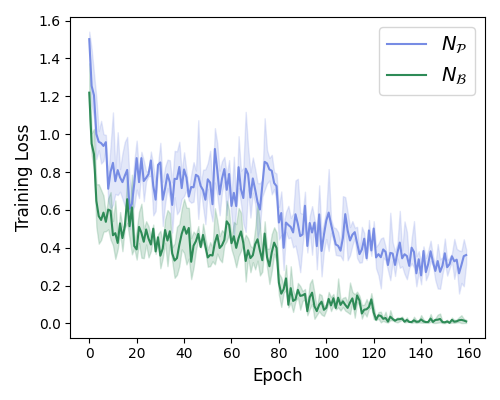}
  \end{center}
  \vspace{-0.8cm}
  \caption{Training loss curves when $N_{\mathcal{P}}$ and $N_{\mathcal{B}}$ are trained on CIFAR 10. The shades of each curve mean the standard deviation of training loss.}
  \vspace{-0.5cm}
  \label{Fig:trend}
\end{wrapfigure}
\textbf{Training Trends.} Simply thinking, it can be expected that the neural network~($N_{\mathcal{P}}$) obtained by pruning will have a similar training tendency to the corresponding base neural network~($N_{\mathcal{B}}$). We checked whether the training tendency was actually similar through a simple experiment. If the training trend is similar, the logit values can be similar and consequently the loss curve will be similar. The loss curves in Fig.\ref{Fig:trend} are obtained using 5 controlled random seeds, and ResNet56 was used as the neural network and CIFAR10 was used as the dataset. In epochs 80 and 120, the learning rate was attenuated by 10 times, and SGD was used as the optimizer, 0.9 for momentum, and 5e-4 for weight decay.\\
\indent As shown in Fig.\ref{Fig:trend}, we can see that the average training loss of $N_{\mathcal{P}}$ is larger over the entire epochs than $N_{\mathcal{B}}$. This means that $N_{\mathcal{P}}$ and $N_{\mathcal{B}}$ have different training trends, that is, logit values from each model are different. However, hyperparameter transferability between $N_{\mathcal{P}}$ and $N_{\mathcal{B}}$ for ResNet56 is high according to Table \ref{tbl:main_results}. Through these two facts, we conclude that it is difficult to measure hyperparameter transferability through loss curve or logit similarity. Therefore, we took a different approach to measure the hyperparameter transferability. \\ \\
\indent \textbf{Structural Similarity.} When the performances of family networks are evaluated in network architecture literatures~\cite{he2016deep,zagoruyko2016wide,huang2017densely}, we noted that hyperparameters do not vary significantly according to the depth or channel width of the neural network. We performed an experiment to verify that the depth of each neural network would not have a significant effect on HPO if each neural network belongs to a family. We performed the following an experiment using ResNet Family~\cite{he2016deep}. First, we acquired hyperparameter sets by performing HPO for ResNet8, ResNet14, and ResNet32 according to Fig.\ref{fig:HPO Procedures}.(a). Next, the performance was measured when the hyperparameter sets obtained earlier were used for ResNet56 training. For comparison, we considered the baseline performance of ResNet56 as the validation accuracy when using the hyperparameter set acquired through Fig.\ref{fig:HPO Procedures} (b), and measured the performance when transferring the hyperparameters of ResNet8, ResNet14, and ResNet32. Each column of Table \ref{tbl:family networks} shows the performance when the hyperparameter set is transferred from each ResNet Family to ResNet56.\\
\begin{table}[!t]
\centering
\caption{Hyperparameter transfer experiments between distinct neural networks. We transferred the hyperparameters obtained from source neural network to target neural network. We measured the accuracies of the validation set. The number in parentheses means the difference between the performance obtained through the proposed HPO framework and the performance when source hyperparameters are transferred to the target neural network.}
\label{tbl:cross_transferability}
\begin{tabular}{c|c|ccc}
\hline
\multirow{2}{*}{$Target$} & \multirow{2}{*}{$Source$}                                                             & \multicolumn{3}{c}{Accuracy (\%)}                                           \\ \cline{3-5} 
                                                                              && \multicolumn{1}{c|}{CIFAR10} & \multicolumn{1}{c|}{CIFAR100} & TinyImageNet \\ \hline
\multirow{3}{*}{VGG16-BN} &
\begin{tabular}[c]{@{}c@{}}VGG16-BN with pruning \end{tabular}    &\multicolumn{1}{c|}{$93.61$}        & \multicolumn{1}{c|}{$73.36$}         & $61.03$              \\ \cline{2-5}
 &
\begin{tabular}[c]{@{}c@{}}RestNet56 \end{tabular}    &\multicolumn{1}{c|}{$93.64(+0.03)$}        & \multicolumn{1}{c|}{$72.31(-1.05)$}         & $61.44(+0.41)$              \\ \cline{2-5}
 &
\begin{tabular}[c]{@{}c@{}}MobileNetV2 \end{tabular}    &\multicolumn{1}{c|}{93.58(-0.03)}        & \multicolumn{1}{c|}{72.54(-0.82)}         & 60.19(-0.84)              \\ \hline
\multirow{3}{*}{MobileNetV2} &
\begin{tabular}[c]{@{}c@{}}MobileNetV2 with pruning \end{tabular}    &\multicolumn{1}{c|}{94.54}        & \multicolumn{1}{c|}{77.89}         & 56.99              \\ \cline{2-5}
 &
\begin{tabular}[c]{@{}c@{}}ResNet56 \end{tabular}    &\multicolumn{1}{c|}{91.95(-2.92)}        & \multicolumn{1}{c|}{75.91(-1.98)}         &55.90(-1.09)              \\ \cline{2-5}
 &
\begin{tabular}[c]{@{}c@{}}VGG16-BN \end{tabular}    &\multicolumn{1}{c|}{94.30(-0.57)}        & \multicolumn{1}{c|}{76.06(-1.83)}         &  56.02(-0.97)              \\ \hline
\multirow{3}{*}{RestNet56} &
\begin{tabular}[c]{@{}c@{}}RestNet56 with pruning \end{tabular}    &\multicolumn{1}{c|}{93.62}        & \multicolumn{1}{c|}{72.31}         &  52.29              \\ \cline{2-5}
 &
\begin{tabular}[c]{@{}c@{}}MobileNetV2 \end{tabular}    &\multicolumn{1}{c|}{93.29(-0.33)}        & \multicolumn{1}{c|}{70.82(-1.49)}         &50.00(-2.29)              \\ \cline{2-5}
 &
\begin{tabular}[c]{@{}c@{}}VGG16-BN \end{tabular}    &\multicolumn{1}{c|}{93.40(-0.22)}        & \multicolumn{1}{c|}{71.77(-0.54)}         &  51.50(-0.79)              \\ \hline
\end{tabular}
\end{table}
\indent As shown in Table \ref{tbl:family networks}, when the hyperparameters obtained from each family network were transferred, there were small differences of around 0.7\%. This result indicates that the hyperparameter set obtained by performing HPO using family neural networks can achieve suboptimal performance. In other words, it means that suboptimal hyperparameter transfer is possible between Family Neural Networks. Based on this observation, we can think of a simple way to effectively reduce the HPO cost. For example, suppose that a neural network~ ($N_{target}$) with high depth among neural network family that can have very small depth like ResNet family is used for application development. In this case, since the dataset to be used is likely not a commonly used dataset, the HPO is inevitable. Depending on the previously observed fact, if we perform HPO using a family neural network~($N_{small}$) with low depth, we can obtain an approximate optimal hyperparameter range. And if we do HPO by allocating a much smaller budget using the acquired hyperparameter range, we will be able to get the optimal hyperparameter of $N_{target}$ in a short time. However, this step-by-step procedure is more inconvenient than Fig.\ref{fig:HPO Procedures}.(b) and has the disadvantage of having to empirically set up the range of each hyperparameter by using the found suboptimal hyperparameters. On the other hand, if HPO is performed through Fig.\ref{fig:HPO Procedures}.(b), optimal  hyperparameter set can be acquired quickly without these drawbacks. \\
\indent From the previous experimental results, we assumed that the hyperparameter transferability was high when the structural similarity was high. In order to support this conjecture, we conducted a hyperparameter transfer experiment between different neural networks rather than family networks~(See Table \ref{tbl:cross_transferability}). The performance of the validation set was measured when the hyperparameter sets obtained by performing HPO using the proxy model of other neural network were used for training of the target neural network. The number in parentheses in each column of Table \ref{tbl:cross_transferability} mean the performance gap that can be obtained through Fig.\ref{fig:HPO Procedures}.(b). The hyperparameter optimizer used in the experiment was TPE~\cite{bergstra2011algorithms}, and the search space pruner was \textit{hyperband}~\cite{li2017hyperband}.\\
\indent From the experimental results, it can be seen that when hyperparameter transfer is performed between VGG16-BN and ResNet56, there is relatively lower performance degradation than the case of hyperparameter transfer between MobileNetV2 and the others~(See Table \ref{tbl:cross_transferability}). On the other hand, looking at the hyperparameter transfer results between MobileNetV2 and the other two neural networks, the performance degradation is relatively high. This trend is consistent across the three datasets. When we look at this trend more closely, the hyperparameter transfer performance gap of ResNet56-MobileNetV2 pair is larger, and VGG16BN-MobileNetV2 pair are smaller than this. Therefore, it can be said that the hyperparameter transferability of VGG16BN-MobileNetV2 pair is higher than that of ResNet56-MobileNetV2 pair. This phenomenon can be attributed to the structural similarity. Consider the basic unit block of VGG16BN and ResNet56. VGG16BN's basic unit block consists of convolution-batchnorm-activation. And the basic unit block of ResNet56 is similar to VGG16BN except an additional convolution-batchnorm-activation block and residual connection. That is, if the two units of the basic block of VGG16BN are connected, there is no difference with the basic block of ResNet56 except for the residual connection. On the other hand, MobileNetV2 not only uses depth-wise convolution, but also the kernel size of the convolution layer is different. Therefore, the pairs including MobileNetV2 showed relatively low hyperparameter transferability. Based on this interpretation for the result of Table \ref{tbl:cross_transferability}, we concluded that it is appropriate to conduct hyperparameter transfer between neural networks with high structural similarity in order to effectively transfer hyperparameters. \\ \\
\begin{figure}[t]
    \centering
    \begin{center}
        \includegraphics[trim=0.0cm 0.0cm 0cm 0.0cm,width=0.8\columnwidth]{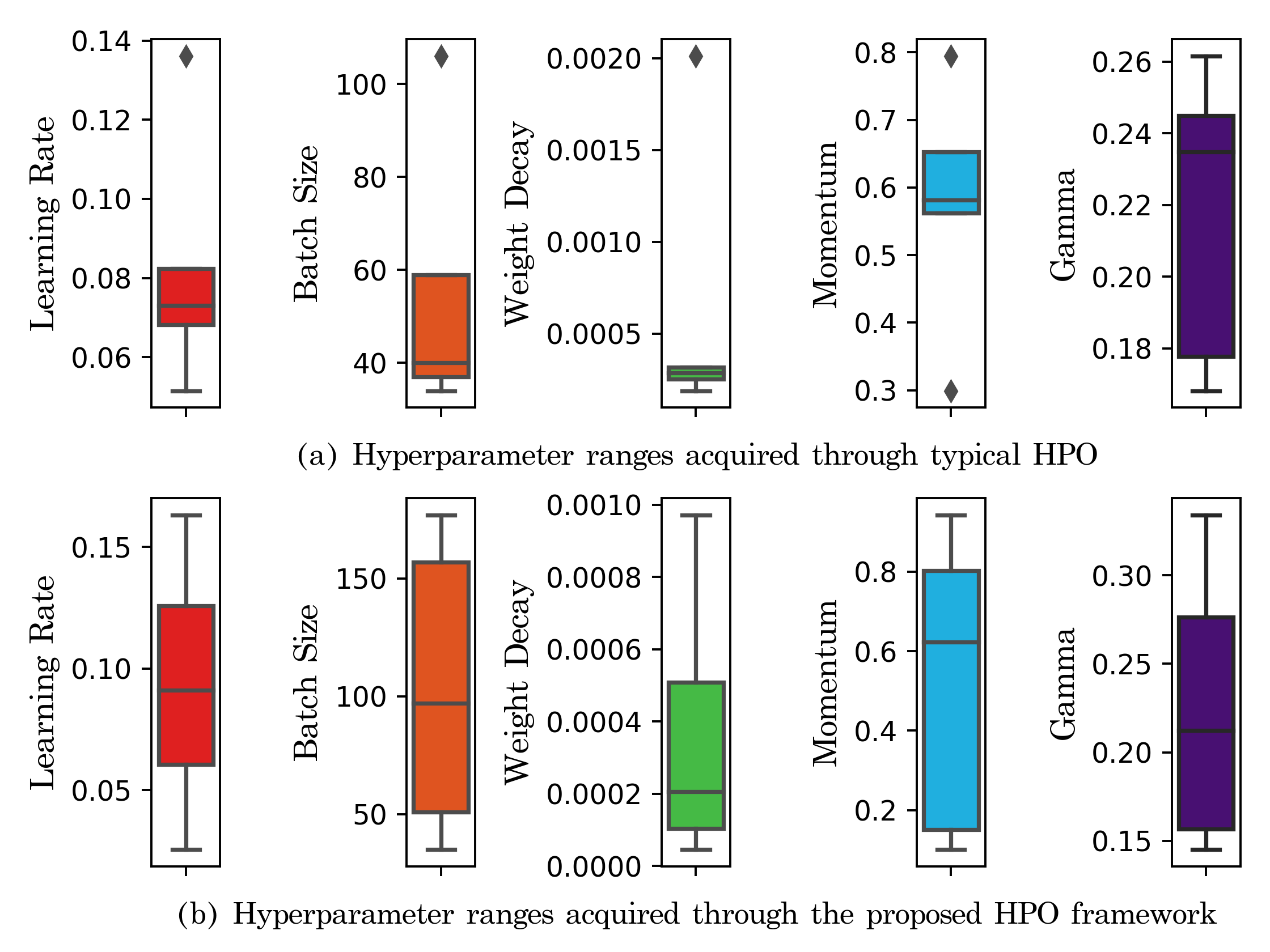}
    \end{center}
    \caption{(a), (b) show the range of hyperparameters acquired from typical HPO and from HPO using proxy model, respectively. ResNet56 and CIFAR100 were used as a model and a dataset. Momentum and Gamma represent the momentum of SGD, and learning-rate-decay-term, respectively. Each horizontal line in a box plot represents the mean value.}
    \label{fig:HP distributions}
\end{figure}
\indent \textbf{Additional hyperparameters.} We confirmed from Table \ref{tbl:main_results} that when the proxy model is used for HPO, the amount of computation can be significantly reduced and comparable performance can be achieved. Extending this result, we performed additional experiment to confirm that our proposal is still valid even if we optimize more hyperparameters. In this additional experiment, we added the momentum and learning rate decay term of the SGD optimizer as new hyperparameters. We used CIFAR100 and ResNet56 as a dataset and neural network, respectively. We used 5 random seeds in the experiment, and unlike the previous experiments, because the number of hyperparameters increased, the trial was increased to 100 times. And we used TPE~\cite{bergstra2011algorithms} and \textit{hyperband}~\cite{li2017hyperband} as hyperparameter optimizer and search space pruner, respectively.\\
\indent Hyperparameters obtained by performing typical HPO showed accuracy for the validation set of $69.83 \pm 1.33$. And the hyperparameters obtained by HPO using $N_{\mathcal{P}}$ as a proxy model showed accuracy for the validation set of $70.57 \pm 1.03$. These results are similar to the trend of Table \ref{tbl:main_results}, showing that our proposal is still valid even if the number of hyperparameters increases. Meanwhile, we compared the ranges of hyperparameters obtained through Fig.\ref{fig:HPO Procedures} (a) and (b). While the learning rate, weight decay, and learning rate decay term have similar mean value, the batch size and momentum show significant differences with respect to mean values~(See \ref{fig:HP distributions}). Although the two hyperparameters show significant difference, the reason why the hyperparameters obtained through each HPO showed similar performance is because the importance of each hyperparameter is different~\cite{zimmer2021auto}. According to zimmer et al.~\cite{zimmer2021auto}, it can be seen that learning rate, learning rate decay term, and weight decay are important hyperparameters in that order, and the other two hyperparameters are relatively less important. In other words, as the hyperparameter has a wide range of values, its importance is lower. As a result, even if we optimize hyperparameters by using our proposal, the results are consistent with the context of previous study~\cite{zimmer2021auto}.

\section{Conclusions}
\label{conclusion}
Despite the advance of hyperparameter optimization~(HPO) methods, the cost of HPO for deep learning models is very high because the computational cost of the deep learning model itself is huge. Inspired by Bayesian optimization using a surrogate model, we propose to use the neural network obtained through neural network pruning as a proxy model for HPO. To verify this idea, we performed extensive experiments using commonly used neural networks and datasets and representative hyperparameter optimization methods. Through these experiments, we demonstrated that using the proxy model for HPO is computationally efficient and can achieve comparable performance. The proposed HPO using proxy model is simple: (1) After pruning a neural network, HPO is performed with the pruned neural network. (2) Then, the best model can be obtained by training the base neural network with the best hyperparameters acquired in the HPO. Therefore, we believe that the proposed method can be easily applied to various existing HPO frameworks while significantly reducing the computational cost.\\
\indent Recently, sustainable AI has received attention from AI society. One of the major factors in which sustainable AI has attracted attention in AI society is the $CO_{2}$ emission. Training a deep learning model generates the tons of $CO_{2}$ emissions~\cite{patterson2021carbon}. This problem can no longer be ignored. For our future generation, many researchers are figuring out to reduce the emission of $CO_{2}$ through various ways. As part of this effort, we hope that our research will help reduce emissions of $CO_{2}$.
%
%
\bibliographystyle{unsrt}
\bibliography{egbib}

\begin{thebibliography}{10}

\bibitem{bergstra2011algorithms}
James Bergstra, R{\'e}mi Bardenet, Yoshua Bengio, and Bal{\'a}zs K{\'e}gl.
\newblock Algorithms for hyper-parameter optimization.
\newblock {\em Advances in neural information processing systems}, 24, 2011.

\bibitem{bergstra2013making}
James Bergstra, Daniel Yamins, and David Cox.
\newblock Making a science of model search: Hyperparameter optimization in
  hundreds of dimensions for vision architectures.
\newblock In {\em International conference on machine learning}, pages
  115--123. PMLR, 2013.

\bibitem{wistuba2015hyperparameter}
Martin Wistuba, Nicolas Schilling, and Lars Schmidt-Thieme.
\newblock Hyperparameter search space pruning--a new component for sequential
  model-based hyperparameter optimization.
\newblock In {\em Joint European Conference on Machine Learning and Knowledge
  Discovery in Databases}, pages 104--119. Springer, 2015.

\bibitem{li2017hyperband}
Lisha Li, Kevin Jamieson, Giulia DeSalvo, Afshin Rostamizadeh, and Ameet
  Talwalkar.
\newblock Hyperband: A novel bandit-based approach to hyperparameter
  optimization.
\newblock {\em The Journal of Machine Learning Research}, 18(1):6765--6816,
  2017.

\bibitem{balandat2020botorch}
Maximilian Balandat, Brian Karrer, Daniel~R. Jiang, Samuel Daulton, Benjamin
  Letham, Andrew~Gordon Wilson, and Eytan Bakshy.
\newblock {BoTorch: A Framework for Efficient Monte-Carlo Bayesian
  Optimization}.
\newblock In {\em Advances in Neural Information Processing Systems 33}, 2020.

\bibitem{ha2019bayesian}
Huong Ha, Santu Rana, Sunil Gupta, Thanh Nguyen, Svetha Venkatesh, et~al.
\newblock Bayesian optimization with unknown search space.
\newblock {\em Advances in Neural Information Processing Systems}, 32, 2019.

\bibitem{kandasamy2020tuning}
Kirthevasan Kandasamy, Karun~Raju Vysyaraju, Willie Neiswanger, Biswajit Paria,
  Christopher~R Collins, Jeff Schneider, Barnabas Poczos, and Eric~P Xing.
\newblock Tuning hyperparameters without grad students: Scalable and robust
  bayesian optimisation with dragonfly.
\newblock {\em J. Mach. Learn. Res.}, 21(81):1--27, 2020.

\bibitem{cakmak2020bayesian}
Sait Cakmak, Raul Astudillo~Marban, Peter Frazier, and Enlu Zhou.
\newblock Bayesian optimization of risk measures.
\newblock {\em Advances in Neural Information Processing Systems},
  33:20130--20141, 2020.

\bibitem{jamieson2016non}
Kevin Jamieson and Ameet Talwalkar.
\newblock Non-stochastic best arm identification and hyperparameter
  optimization.
\newblock In {\em Artificial intelligence and statistics}, pages 240--248.
  PMLR, 2016.

\bibitem{lecun1990optimal}
Yann LeCun, John~S Denker, and Sara~A Solla.
\newblock Optimal brain damage.
\newblock In {\em NIPS}, pages 598--605, 1990.

\bibitem{lee2018snip}
Namhoon Lee, Thalaiyasingam Ajanthan, and Philip~HS Torr.
\newblock Snip: Single-shot network pruning based on connection sensitivity.
\newblock {\em arXiv preprint arXiv:1810.02340}, 2018.

\bibitem{wang2020picking}
Chaoqi Wang, Guodong Zhang, and Roger Grosse.
\newblock Picking winning tickets before training by preserving gradient flow.
\newblock 2020.

\bibitem{tanaka2020pruning}
Hidenori Tanaka, Daniel Kunin, Daniel~L Yamins, and Surya Ganguli.
\newblock Pruning neural networks without any data by iteratively conserving
  synaptic flow.
\newblock {\em Advances in Neural Information Processing Systems}, 33, 2020.

\bibitem{van2020single}
Joost van Amersfoort, Milad Alizadeh, Sebastian Farquhar, Nicholas Lane, and
  Yarin Gal.
\newblock Single shot structured pruning before training.
\newblock {\em arXiv preprint arXiv:2007.00389}, 2020.

\bibitem{snoek2012practical}
Jasper Snoek, Hugo Larochelle, and Ryan~P Adams.
\newblock Practical bayesian optimization of machine learning algorithms.
\newblock {\em Advances in neural information processing systems}, 25, 2012.

\bibitem{springenberg2016bayesian}
Jost~Tobias Springenberg, Aaron Klein, Stefan Falkner, and Frank Hutter.
\newblock Bayesian optimization with robust bayesian neural networks.
\newblock {\em Advances in neural information processing systems},
  29:4134--4142, 2016.

\bibitem{yang2020hyperparameter}
Li~Yang and Abdallah Shami.
\newblock On hyperparameter optimization of machine learning algorithms: Theory
  and practice.
\newblock {\em Neurocomputing}, 415:295--316, 2020.

\bibitem{hansen2016cma}
Nikolaus Hansen.
\newblock The cma evolution strategy: A tutorial.
\newblock {\em arXiv preprint arXiv:1604.00772}, 2016.

\bibitem{kim2020neuron}
Woojeong Kim, Suhyun Kim, Mincheol Park, and Geunseok Jeon.
\newblock Neuron merging: Compensating for pruned neurons.
\newblock {\em Advances in Neural Information Processing Systems}, 33, 2020.

\bibitem{liu2017learning}
Zhuang Liu, Jianguo Li, Zhiqiang Shen, Gao Huang, Shoumeng Yan, and Changshui
  Zhang.
\newblock Learning efficient convolutional networks through network slimming.
\newblock In {\em Proceedings of the IEEE international conference on computer
  vision}, pages 2736--2744, 2017.

\bibitem{huang2018data}
Zehao Huang and Naiyan Wang.
\newblock Data-driven sparse structure selection for deep neural networks.
\newblock In {\em Proceedings of the European conference on computer vision
  (ECCV)}, pages 304--320, 2018.

\bibitem{he2019filter}
Yang He, Ping Liu, Ziwei Wang, Zhilan Hu, and Yi~Yang.
\newblock Filter pruning via geometric median for deep convolutional neural
  networks acceleration.
\newblock In {\em Proceedings of the IEEE/CVF Conference on Computer Vision and
  Pattern Recognition}, pages 4340--4349, 2019.

\bibitem{wang2019eigendamage}
Chaoqi Wang, Roger Grosse, Sanja Fidler, and Guodong Zhang.
\newblock Eigendamage: Structured pruning in the kronecker-factored eigenbasis.
\newblock In {\em International Conference on Machine Learning}, pages
  6566--6575. PMLR, 2019.

\bibitem{meng2020pruning}
Fanxu Meng, Hao Cheng, Ke~Li, Huixiang Luo, Xiaowei Guo, Guangming Lu, and Xing
  Sun.
\newblock Pruning filter in filter.
\newblock {\em arXiv preprint arXiv:2009.14410}, 2020.

\bibitem{han2016eie}
Song Han, Xingyu Liu, Huizi Mao, Jing Pu, Ardavan Pedram, Mark~A Horowitz, and
  William~J Dally.
\newblock Eie: Efficient inference engine on compressed deep neural network.
\newblock {\em ACM SIGARCH Computer Architecture News}, 44(3):243--254, 2016.

\bibitem{krizhevsky2009learning}
Alex Krizhevsky, Geoffrey Hinton, et~al.
\newblock Learning multiple layers of features from tiny images.
\newblock 2009.

\bibitem{Le2015TinyIV}
Ya~Le and Xuan Yang.
\newblock Tiny imagenet visual recognition challenge.
\newblock 2015.

\bibitem{simonyan2014very}
Karen Simonyan and Andrew Zisserman.
\newblock Very deep convolutional networks for large-scale image recognition.
\newblock {\em arXiv preprint arXiv:1409.1556}, 2014.

\bibitem{he2016deep}
Kaiming He, Xiangyu Zhang, Shaoqing Ren, and Jian Sun.
\newblock Deep residual learning for image recognition.
\newblock In {\em Proceedings of the IEEE conference on computer vision and
  pattern recognition}, pages 770--778, 2016.

\bibitem{sandler2018mobilenetv2}
Mark Sandler, Andrew Howard, Menglong Zhu, Andrey Zhmoginov, and Liang-Chieh
  Chen.
\newblock Mobilenetv2: Inverted residuals and linear bottlenecks.
\newblock In {\em Proceedings of the IEEE conference on computer vision and
  pattern recognition}, pages 4510--4520, 2018.

\bibitem{loshchilov2016cma}
Ilya Loshchilov and Frank Hutter.
\newblock Cma-es for hyperparameter optimization of deep neural networks.
\newblock {\em arXiv preprint arXiv:1604.07269}, 2016.

\bibitem{optuna_2019}
Takuya Akiba, Shotaro Sano, Toshihiko Yanase, Takeru Ohta, and Masanori Koyama.
\newblock Optuna: A next-generation hyperparameter optimization framework.
\newblock In {\em Proceedings of the 25rd {ACM} {SIGKDD} International
  Conference on Knowledge Discovery and Data Mining}, 2019.

\bibitem{he2015delving}
Kaiming He, Xiangyu Zhang, Shaoqing Ren, and Jian Sun.
\newblock Delving deep into rectifiers: Surpassing human-level performance on
  imagenet classification.
\newblock In {\em Proceedings of the IEEE international conference on computer
  vision}, pages 1026--1034, 2015.

\bibitem{glorot2010understanding}
Xavier Glorot and Yoshua Bengio.
\newblock Understanding the difficulty of training deep feedforward neural
  networks.
\newblock In {\em Proceedings of the thirteenth international conference on
  artificial intelligence and statistics}, pages 249--256. JMLR Workshop and
  Conference Proceedings, 2010.

\bibitem{zagoruyko2016wide}
Sergey Zagoruyko and Nikos Komodakis.
\newblock Wide residual networks.
\newblock {\em arXiv preprint arXiv:1605.07146}, 2016.

\bibitem{huang2017densely}
Gao Huang, Zhuang Liu, Laurens Van Der~Maaten, and Kilian~Q Weinberger.
\newblock Densely connected convolutional networks.
\newblock In {\em Proceedings of the IEEE conference on computer vision and
  pattern recognition}, pages 4700--4708, 2017.

\bibitem{zimmer2021auto}
Lucas Zimmer, Marius Lindauer, and Frank Hutter.
\newblock Auto-pytorch: Multi-fidelity metalearning for efficient and robust
  autodl.
\newblock {\em IEEE Transactions on Pattern Analysis and Machine Intelligence},
  2021.

\bibitem{patterson2021carbon}
David Patterson, Joseph Gonzalez, Quoc Le, Chen Liang, Lluis-Miquel Munguia,
  Daniel Rothchild, David So, Maud Texier, and Jeff Dean.
\newblock Carbon emissions and large neural network training.
\newblock {\em arXiv preprint arXiv:2104.10350}, 2021.

\end{thebibliography}
\end{document}